%% file: main-5947-Nortje.tex
\documentclass[11pt,a4paper]{article}

\usepackage{times,latexsym}
\usepackage{url}
\usepackage[T1]{fontenc}
\usepackage{graphicx}
\usepackage{amssymb,amsmath,bm}
\usepackage{caption}
\usepackage{subcaption}
\usepackage{soul}

\usepackage[acceptedWithA]{tacl2021v1}

\usepackage[utf8]{inputenc} %
\usepackage[T1]{fontenc}    %
\usepackage{hyperref}       %
\usepackage{url}            %
\usepackage{booktabs}       %
\usepackage{amsfonts}       %
\usepackage{nicefrac}       %
\usepackage{microtype}      %
\usepackage{hyperref}
\usepackage{multirow}
\usepackage{xcolor}         %
\usepackage{pifont}
\usepackage{tipa}

\usepackage{booktabs}
\usepackage{tabularx}
\usepackage{multirow}

\newcolumntype{C}{>{\centering\arraybackslash}X}
\newcolumntype{L}{>{\raggedright\arraybackslash}X}
\newcolumntype{R}{>{\raggedleft\arraybackslash}X}
\newcolumntype{P}[1]{>{\raggedright\arraybackslash}p{#1}}

\newcommand{\model}{{\mdseries \scshape Matt\-Net}}

\newcommand{\me}{familiar--\underline{novel}}
\newcommand{\Me}{Familiar--\underline{novel}}
\newcommand{\meOther}{familiar--\underline{novel}$^*$}
\newcommand{\MeOther}{Familiar--\underline{novel}$^*$}
\newcommand{\familiar}{familiar--\underline{familiar}}
\newcommand{\Familiar}{Familiar--\underline{familiar}}
\newcommand{\familiarWithNovel}{\underline{familiar}--novel}
\newcommand{\FamiliarWithNovel}{\underline{Familiar}--novel}
\newcommand{\novel}{novel--\underline{novel}}
\newcommand{\Novel}{Novel--\underline{novel}}

\newcommand{\image}[1]{{\scshape #1}}
\newcommand{\word}[1]{\textit{#1}}

\usepackage{xspace,mfirstuc,tabulary}

\newif\iftaclinstructions
\taclinstructionsfalse %
\iftaclinstructions
\renewcommand{\confidential}{}
\renewcommand{\anonsubtext}{(No author info supplied here, for consistency with
TACL-submission anonymization requirements)}
\newcommand{\instr}
\fi

\iftaclpubformat %

\else

\fi

\definecolor{mycolor}{HTML}{008000}%
\definecolor{indiagreen}{HTML}{138808}%
\definecolor{papaya}{HTML}{EE892F}%
\definecolor{mygreen}{HTML}{008000}%
\definecolor{mypurple}{HTML}{9966CC}%
\definecolor{myblue}{HTML}{5D8AA8}
\definecolor{mypink}{HTML}{EC008C}

\newcommand{\changed}[1]{#1}%

\usepackage{pdflscape}

\newcommand{\first}{$^\ast$}
\newcommand{\second}{$^\blacklozenge$}
\newcommand{\third}{$^\bullet$}

\newcommand{\ab}{\mathbf{a}}
\newcommand{\vb}{\mathbf{v}}

\title{Visually Grounded Speech Models have a Mutual Exclusivity Bias}

\author{
  Leanne Nortje\first \quad
  Dan Oneață\second \quad
  Yevgen Matusevych\third \quad
  Herman Kamper\first \quad
  \\
  \\
  \first Electrical and Electronic Engineering, Stellenbosch University, South Africa
  \\
  \second SpeeD Lab, University Politehnica of Bucharest, Romania
  \\
  \third CLCG, University of Groningen, the Netherlands
  \\
  {\small \texttt{nortjeleanne@gmail.com} \quad \texttt{dan.oneata@gmail.com}} \\
  {\small \texttt{yevgen.matusevych@rug.nl} \quad \texttt{kamperh@sun.ac.za}}  
}

\date{}
\begin{document}
    \maketitle
    
    \begin{abstract}
    When children learn new words, they employ constraints such as the mutual exclusivity (ME) bias: a novel word is mapped to a novel object rather than a familiar one.
    This bias has been studied computationally, but only in models that use discrete word representations as input, ignoring the high variability of spoken words. 
    We investigate the ME bias in the context of visually grounded speech models that learn from natural images and continuous speech audio.
    Concretely, we train a model on familiar words and test its ME bias by asking it to select between a novel and a familiar object when queried with a novel word.
    To simulate prior acoustic and visual knowledge, we experiment with several initialisation strategies using pretrained speech and vision networks.
    Our findings reveal the ME bias across the different initialisation approaches, with a stronger bias in models with more prior (in particular, visual) knowledge.
    Additional tests confirm the robustness of our results, %
    \changed{even when different loss functions are considered. Based on detailed analyses to piece out the model's representation space, we attribute
    the ME bias to how familiar and novel classes are distinctly separated in the resulting space.}
    \end{abstract}

    \input{introduction}
    \input{related-work}
    \input{task}
    \input{model}
    \input{results}
    \input{further-analysis}
    \input{conclusion}

    \bibliography{mybib.bib}
    \bibliographystyle{acl_natbib}
    
    \appendix
    \input{more_results}

\end{document}

%% file: introduction.tex
\section{Introduction}
\label{sec:intro}

When children learn new words, they employ a set of basic constraints to make the task easier. 
One such constraint is the \textit{mutual exclusivity} (ME) bias: when a learner hears a novel word, they map it to an unfamiliar object (whose name they don't know yet), rather than a familiar one. 
This strategy was first described by \citet{markman_childrens_1988} over 30 years ago and has since been studied extensively in the developmental sciences~\cite{merriman_mutual_1989, markman_use_2003, mather_learning_2009, lewis_role_2020}. 
With the rise of neural architectures, recent years saw renewed interest in the ME bias, this time from the computational modelling perspective: several studies have examined whether and under which conditions the ME bias emerges in machine learning models \citep{gulordava_deep_2020, gandhi_mutual_2020, vong_cross-situational_2022, ohmer_mutual_2022}.

The models in these studies normally receive input consisting of word and object representations, as the ME strategy is used to learn %
mappings between words and the objects they refer to. 
Object representations vary in their complexity, from symbolic representations of single objects~\citep[e.g.,][]{gandhi_mutual_2020} to continuous vectors encoding a natural image~\citep[e.g.,][]{vong_cross-situational_2022}.  
Word representations, however, are based on their written forms in all these studies.
E.g., the textual form of the word \word{fish} has an invariable representation in the input.
This is problematic because children learn words from continuous speech, 
and there is large variation in how the word \word{fish} can be realised depending on the word duration, prosody, the quality of the individual sounds and so on; see e.g.,~\citet{creel_phonological_2012} on how the ME bias affects atypical pronunciations such as \textipa{[fesh]} instead of \textipa{[fIsh]}. 
As a result, children face an additional challenge compared to models trained on written words. This is why it is crucial to investigate the ME bias in a more naturalistic setting, with models trained on word representations that take into account variation between acoustic instances of the same word. 

Recently, there has been a lot of headway in the development of visually grounded speech models that learn from images paired with unlabelled speech~\cite{harwath_unsupervised_2016, harwath_vision_2018, kamper_semantic_2019, chrupala_visually_2022, peng_fast-slow_2022, peng_syllable_2023, berry_m-speechclip_2023, shih_speechclip_2023}.
Several studies have shown, for instance, that these models learn word-like units when trained on large amounts of paired speech--vision data~\cite{harwath_learning_2017,harwath_jointly_2018, olaleye_keyword_2022, peng_word_2022, nortje_towards_2023, pasad_what_2023}.
Moreover, some of these models draw inspiration from the way infants acquire language from spoken words that co-occur with visual cues across different situations in their environments~\cite{miller_how_1987, yu_rapid_2007, cunillera_speech_2010, thiessen_effects_2010}. However, the ME bias has not been studied in these models.
 
In this work we test whether visually grounded speech models exhibit the ME bias. We focus on a recent model by \citet{nortje_visually_2023-1},
as it achieves state-of-the-art performance in a few-shot learning task that resembles the word learning setting considered here. The model's architecture is representative of many of the other recent visually grounded speech models: it takes a spoken word and an image as input, processes these independently, and then relies on a word-to-image attention mechanism to learn a mapping between a spoken word and its visual depiction.
We first train the model to discriminate familiar words.
We then test its ME bias by presenting it with a novel word and two objects, one familiar and one novel. 
To simulate prior acoustic and visual knowledge that a child might have already acquired before word learning, we additionally explore different initialisation strategies for the audio and vision branches of the model.

To preview our results, we observe the ME bias across all the different initialisation schemes of the visually grounded speech model, and the bias is stronger in models with more prior \changed{visual} knowledge. We also carry out a series of additional tests to ensure that the observed ME bias is not merely an artefact, and present analyses to pinpoint the relationship between the model's representation space and the emergence of the ME bias.
\changed{In experiments where we look at different modelling options (visual initialisation and loss functions, in particular), the ME bias is observed in all cases.}
\changed{The code and the accompanying dataset are available from our project website.\footnote{\tiny \url{https://sites.google.com/view/mutualexclusivityinvgs}}}

%% file: related-work.tex
\section{Related work}
\label{sec:related_work}

Visually grounded speech models learn by bringing together representations of paired images and speech while pushing mismatched pairs apart.
These models have been used in several downstream tasks, ranging from speech--image retrieval \cite{harwath_jointly_2018} and keyword spotting~\cite{olaleye_keyword_2022} to word~\cite{peng_word_2022} and syllable segmentation~\cite{peng_syllable_2023}.

\changed{In terms of design choices, early models used a hinge loss~\cite{harwath_unsupervised_2016, harwath_jointly_2018},
while several more advanced losses have been proposed since~\cite{petridis_end--end_2018,peng_fast-slow_2022, peng_syllable_2023}.
A common strategy to improve performance is to initialise the vision branch using a supervised vision model, e.g., \citet{harwath_unsupervised_2016} used VGG, \citet{harwath_learning_2020} used ResNet, and recently \citet{shih_speechclip_2023} and \citet{berry_m-speechclip_2023} used CLIP.
For the speech branch, self-supervised speech models like wav2vec2.0 and HuBERT have been used for initialisation~\cite{peng_word_2022}.
Other extensions include using vector quantisation in intermediate layers~\cite{harwath_learning_2020} and more advanced multimodal attention mechanisms to connect the branches~\cite{chrupala_representations_2017, radford_learning_2021, peng_fast-slow_2022, peng_self-supervised_2022}.} 

\changed{In this work we specifically use the few-shot model of~\citet{nortje_visually_2023-1} that incorporates many of these strategies (Section~\ref{sec:visually_grounded_model}). We also look at how different design choices affect our analysis of the ME bias, e.g., using different losses (Section~\ref{sec:general_experiments}).}

\changed{As noted already, previous computational studies of the ME bias have exclusively used the written form of words as input~\cite{gulordava_deep_2020, gandhi_mutual_2020, vong_cross-situational_2022, ohmer_mutual_2022}.
Visually grounded speech models have the benefit that they can take} %
real speech as input. %
This better resembles the actual experimental setup with human participants~\cite{markman_childrens_1988,markman_categorization_1989}.

\changed{Concretely,}
since the models in \citet{gulordava_deep_2020} and \citet{vong_cross-situational_2022} are trained on written words, which are discrete by design, they need to learn a continuous embedding for each of the input classes.
However, this makes dealing with novel inputs difficult: if a model never sees a particular item at training time, its embeddings are never updated and remain randomly initialised. 
As a result, the ME test becomes a comparison of learned vs random embeddings instead of novel vs familiar.
To address this issue, \citet{gulordava_deep_2020} use novel examples in their contrastive loss during training, while \citet{vong_cross-situational_2022} perform one gradient update on novel classes before testing.
These strategies mean that, in both cases, the learner has actually seen the novel classes before testing.
Such adaptations are necessary in models taking in written input. 
In contrast, a visually grounded speech model, even when presented with an arbitrary input sequence, can place it in the representation space learned from the familiar classes during training.
\changed{We investigate whether such a representation space results in the ME bias.}

%% file: task.tex
\section{Mutual exclusivity in visually grounded speech models}
\label{sec:me}

\begin{figure}[!t]
	\centering
	\includegraphics[width=\linewidth]{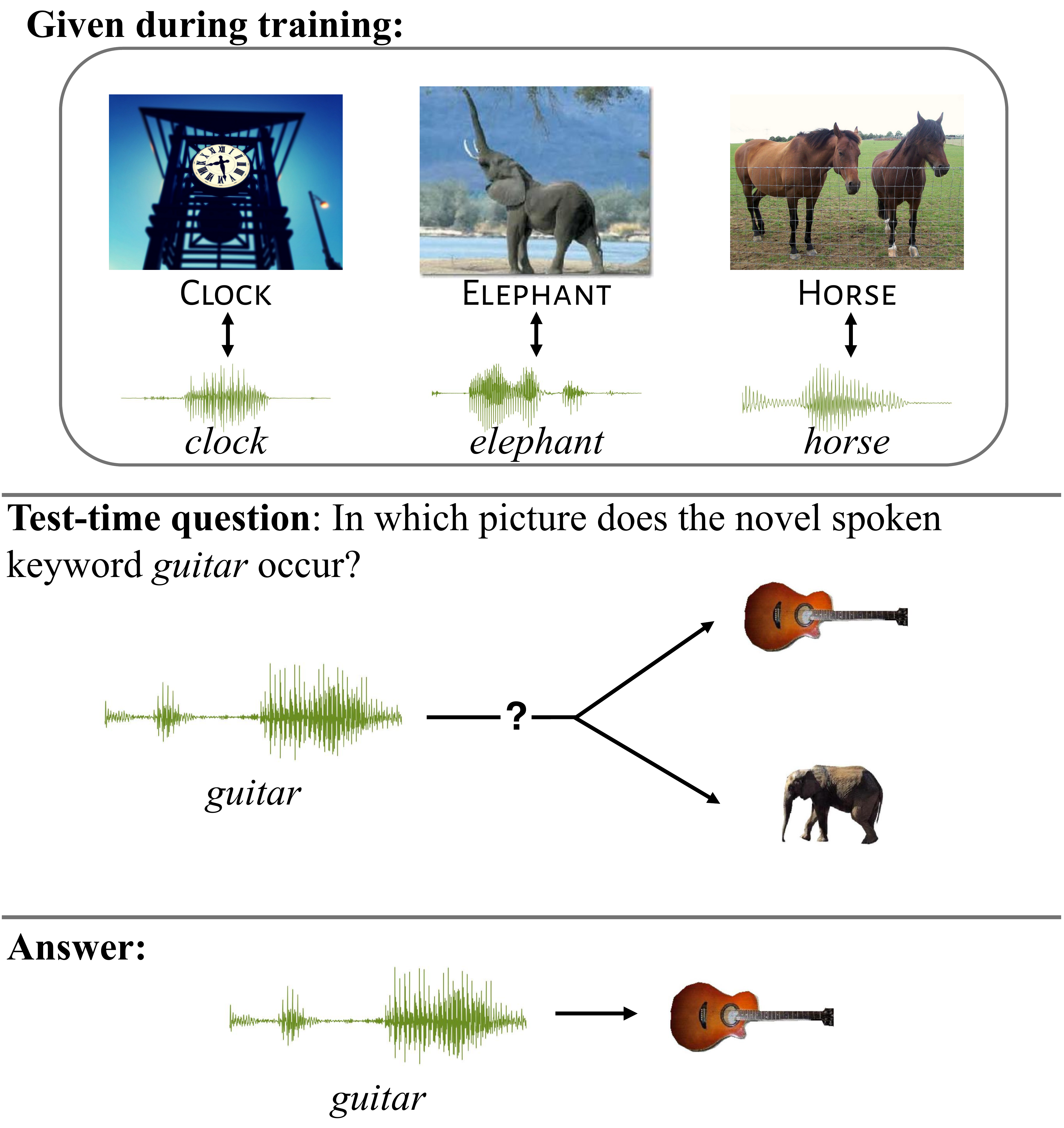}
    \caption{
        \textit{Top:} A learner is familiarised with a set of objects during training.
        \textit{Middle:} At test time, two images are given, one from a familiar class seen during training and the other from an unseen novel class.
        \textit{Bottom:} If a learner has a ME bias, then when prompted with a novel spoken query, the novel object (\image{guitar}) would be selected.}
	\label{fig:task}
\end{figure}

Mutual exclusivity (ME) is a constraint used to learn words.
It is grounded in the assumption that an object, once named, cannot have another name.
The typical setup of a ME experiment \citep{markman_childrens_1988} involves two steps and is illustrated in Figure~\ref{fig:task}.
First, 
the experimenter will
ensure that
the learner (usually a child)
is familiar with a set of specific objects %
by assessing 
their ability to correctly identify objects associated with familiar words.
In this example, the familiar classes are {`clock'}, {`elephant'} and {`horse'}, as illustrated in the top panel of the figure.
Subsequently, at test time the learner is shown a familiar image (e.g., \image{elephant}) and a novel image (e.g., \image{guitar})
and is asked to determine which of the two corresponds to a novel spoken word, e.g., \word{guitar} (middle panel in the figure).
If the learner exhibits a ME bias, they would select the corresponding novel object, \image{guitar} in this case (bottom panel).

Our primary objective is to investigate the ME bias in computational models that operate on the audio and visual modalities.
These models, known as visually grounded speech models, draw inspiration from how children learn words~\cite{miller_how_1987}, by being trained on unlabelled spoken utterances paired with corresponding images. The models learn to associate spoken words and visual concepts, and often do so by predicting a similarity score for a given audio utterance and an input image.
This score can then be used to select between competing visual objects given a spoken utterance, as required in the ME test.

In Section~\ref{sec:data_setup} below we describe how we set up our test of the ME bias.
In Section~\ref{sec:visually_grounded_model} we then present the visually grounded speech model that we use in this study.

\section{Constructing a speech--image test for mutual exclusivity}
\label{sec:data_setup}

To construct our ME test, we need isolated spoken words that are paired with natural images of objects. We also need to separate these paired word--image instances into two sets: familiar classes and novel classes. A large multimodal dataset of this type does not exist, so we create one by combining several image and speech datasets.

For the images, we combine MS\,COCO~\cite{lin_microsoft_2014} and Caltech-101~\cite{fei-fei_one-shot_2006}.
MS\,COCO contains 328k images of 91 objects in their natural environment.
Caltech-101 contains 9k %
Google images spanning 101 classes.
Ground truth object segmentations are available for both these datasets.
During training, we use entire images, but during evaluation, we use segmented objects.
This resembles a naturalistic learning scenario in which a learner is familiarised with objects by seeing them in a natural context, but is presented with individual objects (or their pictures) in isolation at test time.

For the audio, we combine the FAAC~\cite{harwath_deep_2015}, Buckeye~\cite{pitt_mark_buckeye_2005} and LibriSpeech~\cite{panayotov_librispeech_2015} datasets.
These English corpora respectively span 183, 40 and 2.5k speakers. 

\begin{table}[bt]
    \small
    \centering
    \renewcommand{\arraystretch}{1.2}
    \begin{tabularx}{\linewidth}{@{}lL@{}}
        \toprule
        Familiar & bear, bird, boat, car, cat, clock, cow, dog, elephant, horse, scissors, sheep, umbrella \\
        Novel & ball, barrel, bench, buck, bus, butterfly, cake, camera, canon, chair, cup, fan, fork, guitar, lamp, nautilus, piano, revolver, toilet, trumpet \\
        \bottomrule
    \end{tabularx}
    \caption{The familiar and novel classes in our ME test setup.}
    \label{tab:classes}
\end{table}

To select familiar and novel classes, we do a manual inspection to make sure that object segmentations for particular classes are of a reasonably high quality
and that there are enough spoken instances for each class in the segmented speech data (at least 100 spoken examples per class).
As an example of an excluded class, we did not use \image{curtain}, since it was often difficult to reliably see that curtains are depicted after these are segmented out.
The final result is a setup with 13 familiar classes and 20 novel classes, as listed in Table~\ref{tab:classes}.

During training (Figure~\ref{fig:task}, top panel), a model only sees familiar classes. 
We divide our data so that we have a training set with 18,279 unique spoken word segments and 94,316 unique unsegmented natural images spanning the 13 familiar classes. 
These are then paired up for training as explained in Section~\ref{sec:model}. 
During training we also use a development set for early stopping; this small set consists of 130 word segments and 130 images from familiar classes.

For ME testing (Figure~\ref{fig:task}, middle panel) we require a combination of familiar and novel classes. 
Our test set in total consists of 8,575 spoken word segments with 22,062 segmented object images. 
To implement the ME test, we sample 1k episodes: each episode consists of a novel spoken word (query) with two sampled images, one matching the novel class from the query and the other containing a familiar object.
We ensure that the two images always come from the same image dataset to avoid any intrinsic dataset biases.
There is no overlap between training, development and test samples.

%% file: model.tex
\section{A visually grounded speech model}
\label{sec:visually_grounded_model}

\begin{figure}[!b]
	\centering
	\includegraphics[width=0.9\linewidth]{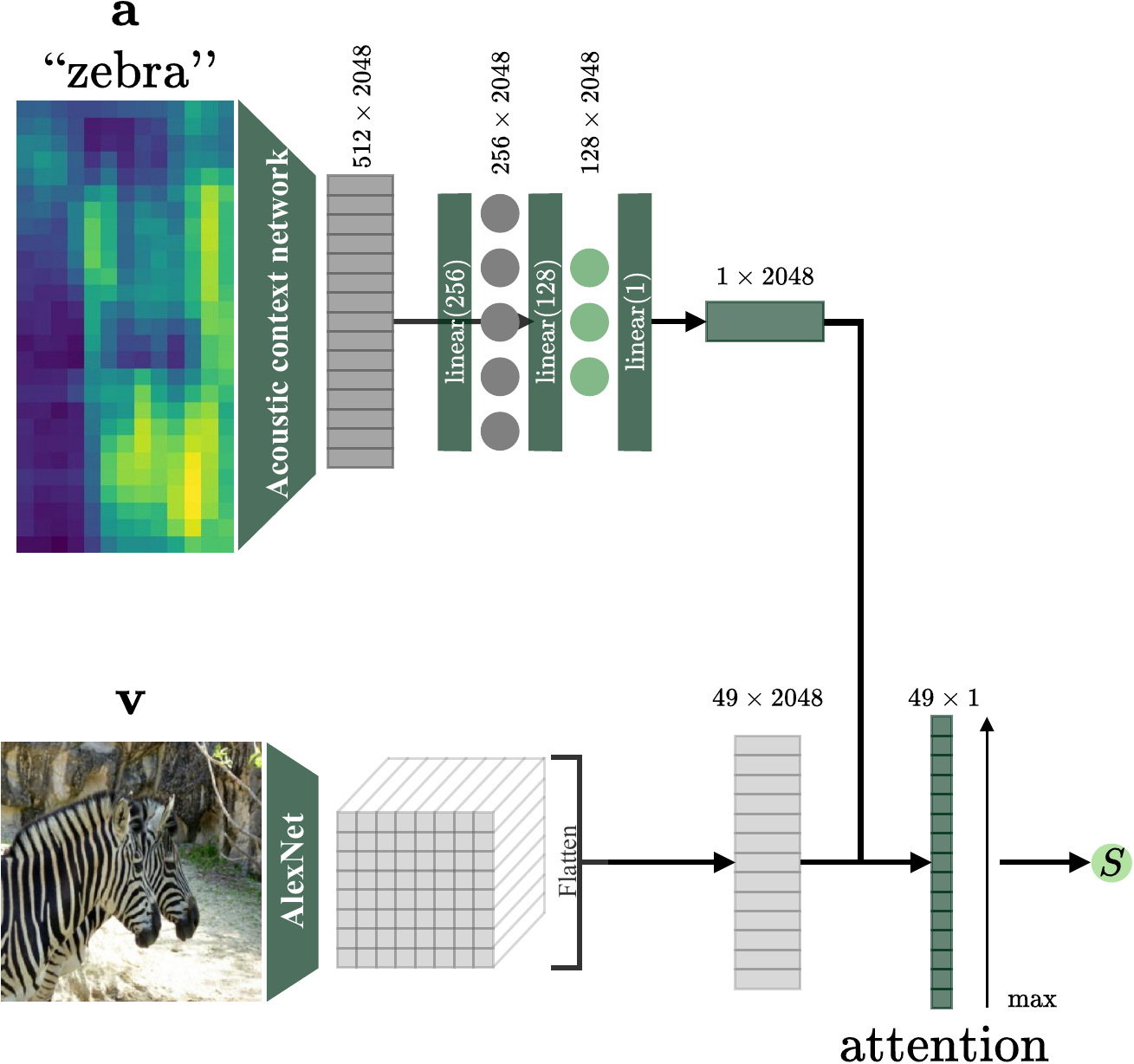}
	\caption{%
        \model~\cite{nortje_visually_2023-1} consists of a vision network and an audio network. These are connected through a word-to-image attention mechanism. The model outputs a score $S$ indicating the similarity of the speech and image inputs. 
    }
	\label{fig:model}
\end{figure}

We want to establish whether visually grounded speech models exhibit the ME bias.
While there is a growing number of speech--image models (Section~\ref{sec:related_work}), many of them share the same general methodology.
We therefore use a visually grounded speech model that is representative of the models in this research area: the \textsc{M}ultimodal \textsc{att}ention \textsc{Net}work (\model) of \citet{nortje_visually_2023-1}.
This model achieves top performance in a few-shot word--object learning task that resembles the way infants learn words from limited exposure. 
Most useful for us is that the model is conceptually simple: it takes an image and a spoken word and outputs a score indicating how similar the inputs are, precisely what is required for ME testing. %

\subsection{Model}
\label{sec:model}

\model\ consists of a vision and an audio branch that are connected with a word-to-image attention mechanism, as illustrated in Figure~\ref{fig:model}.

A spoken word $\mathbf{a}$ is first parameterised as a mel-spectrogram with a hop length of 10 ms, a window of 25 ms and 40 bins.
The audio branch takes this input,
passes it through an acoustic network consisting of LSTM and BiLSTM layers, and finally outputs a single word embedding by pooling the sequence of representations along the time dimension with a two-layer feedforward network.
This method of encoding a variable-length speech segment into a single embedding is similar to the idea behind acoustic word embeddings~\cite{chung_unsupervised_2016, holzenberger_learning_2018, wang_segmental_2018, kamper_truly_2019}.

The vision branch is an adaptation of AlexNet~\cite{krizhevsky_imagenet_2017}.
An image $\mathbf{v}$ is first resized to 224$\times$224 pixels and normalised with means and variances calculated on ImageNet~\cite{deng_imagenet_2009}.
The vision branch then encodes the input image into a sequence of pixel embeddings.

The audio and vision branches are connected through a multimodal attention mechanism that takes the dot product between the acoustic word embedding and each pixel embedding. The maximum of these attention scores is taken as the final output of the model, the similarity score $S$. The idea behind this attention mechanism is to focus on the regions within the image that are most indicative of the spoken word.

The similarity score $S(\mathbf{a}, \mathbf{v})$ should be high if the spoken word $\mathbf{a}$ and the image $\mathbf{v}$ are instances of the same class, and low otherwise.
This is accomplished by using a contrastive loss that pushes positive word--image pairs from the same class closer together than mismatched negative word--image pairs~\cite{nortje_visually_2023-1}:
\begin{equation}
    \small
    \begin{aligned}	
        \ell = &\ d\left(S(\boldsymbol{\mathrm{a}}, \boldsymbol{\mathrm{v}}), 100\right)\\	
        &+\sum_{i=1}^{N_\textrm{neg}}d\left(S(\boldsymbol{\mathrm{a}}^{-}_{i}, \boldsymbol{\mathrm{v}}), 0\right) + \sum_{i=1}^{N_\textrm{neg}}d\left(S(\boldsymbol{\mathrm{a}}, \boldsymbol{\mathrm{v}}^{-}_{i}), 0\right)\\
        &+ \sum_{i=1}^{N_\textrm{pos}}d\left(S(\boldsymbol{\mathrm{a}}, \boldsymbol{\mathrm{v}}_{i}^+), 100\right) + \sum_{i=1}^{N_\textrm{pos}}d\left(S(\boldsymbol{\mathrm{a}}_{i}^+, \boldsymbol{\mathrm{v}}), 100\right)
        \label{eq:loss}
    \end{aligned}
\end{equation}%
\noindent where
$d$ is the squared Euclidean distance, i.e., $S$ is pushed to 0 for negative pairs and to 100 for positive pairs.
In more detail, for an anchor positive word--image pair $(\mathbf{a}, \mathbf{v})$, we sample positive examples ($\boldsymbol{\mathrm{a}}_{1:N_{\textrm{pos}}}^+$, $\boldsymbol{\mathrm{v}}_{1:N_{\textrm{pos}}}^+$) that match the class of the anchor and negative examples ($\boldsymbol{\mathrm{a}}^{-}_{1:N_{\textrm{neg}}}$, $\boldsymbol{\mathrm{v}}^{-}_{1:N_{\textrm{neg}}}$) that are not instances of the anchor class.
We use 
$N_\textrm{pos}=5$ and $N_\textrm{neg}=11$ in our implementation.

As a reminder from Section~\ref{sec:data_setup}, the model is trained exclusively on familiar classes and never sees any novel classes during training. 
Novel classes are also never used as negative examples. 
We train the model with Adam~\cite{kingma_adam_2015} for 100 epochs and use early stopping with a validation task. 
The validation task involves presenting the model with a familiar word query and asking it to identify which of the two familiar object images it refers to. 
We use the spoken words and isolated object images from the development set for this task (see Section~\ref{sec:data_setup}).

\subsection{Different initialisation strategies as a proxy for prior knowledge}
\label{sec:initialisations}

\begin{table*}[!t]
        \small
     	\newcommand{\cmark}{\textcolor{black}{\ding{51}}}%
        \newcommand{\xmark}{\textcolor{gray!30}{\ding{55}}}%
	\centering
	\renewcommand{\arraystretch}{1.2}
        \newcommand{\ii}[1]{{\footnotesize \textcolor{gray}{#1}}} 
	\begin{tabularx}{\linewidth}
    {@{}rlCCCC@{}}
		\toprule
            & & \multicolumn{2}{c}{Model initialisation} & 
		\multicolumn{2}{c}{Accuracy (\%)}\\
		\cmidrule(rl){3-4}
		\cmidrule(l){5-6}
		& & Audio (CPC) & Vision (AlexNet) & \Familiar & \Me\\
		\midrule
		\ii{1} & Random baseline & N/A & N/A & 50.19 & 49.92\\
            \addlinespace
            \ii{2} & \multirow{4}{*}{\model} & \xmark & \xmark & 72.86 & 57.29 \\
            \ii{3} &                         & \xmark & \cmark & 85.89 & 59.32 \\
            \ii{4} &                         & \cmark & \xmark & 75.78 & 55.92 \\
            \ii{5} &                         & \cmark & \cmark & 83.20 & 60.27 \\
		\bottomrule
	\end{tabularx}
        \caption{Performance for different initialisation strategies of \model. The ME results are given in the \me\ column. As a reference, discrimination performance between familiar classes is given under \familiar. }
	\label{tab:me}
\end{table*}

The ME bias has been observed in children at the age of around 17 months \citep[e.g.,][]{halberda_development_2003}.
At this age, children have already gained valuable experience from both spoken language used in their surroundings and the visual environment that they navigate~\cite{clark_how_2004}. For example, 4.5-month-olds can recognise objects \citep{needham_object_2001}, and 6.5-month-olds can recognise some spoken word forms \citep{jusczyk_infants_1995}.
These abilities can be useful when learning new words. 
In light of this, we adopt an approach that initialises the vision and audio branches of our model to emulate prior knowledge.

For the vision branch, we use the convolutional encoder of the \changed{self-supervised AlexNet~\cite{koohpayegani_compress_2020}, which distils the SimCLR ResNet50x4 model \cite{chen_simple_2020} into AlexNet and trains it on ImageNet~\cite{deng_imagenet_2009}.}
For the audio branch, we use an acoustic network~\cite{van_niekerk_vector-quantized_2020} pretrained on the LibriSpeech~\cite{panayotov_librispeech_2015} and Places~\cite{harwath_vision_2018} datasets using a self-supervised contrastive predictive coding (CPC) objective~\cite{oord_representation_2019}. 
\changed{Both these initialisation networks are trained without supervision directly on unlabelled speech or vision data, again emulating the type of data an infant would be exposed to.}
When these initialisation strategies are not in use, we initialise the respective branches randomly.

Considering these strategies, we end up with four possible \mbox{\model}\ variations: one where both the vision and audio branches are initialised from pretrained networks, one where only the audio branch is initialised from a CPC model, one where only the vision branch is initialised from AlexNet, and one where neither branch is initialised with pretrained models (i.e., a full random initialisation). 

In the following sections, we present our results. 
We compare them to the performance of a naive baseline which chooses one of the two images at random for a given word query.
To determine whether the differences between our model variations and a random baseline are statistically significant, we fit mixed-effects regression models to \model's scores using the lme4 package \citep{bates_fitting_2015}. Details are given in Appendix~\ref{append:stats}. 

%% file: results.tex
\section{Mutual exclusivity results}
\label{sec:results}

Our main question is whether visually grounded models like \model\ (Section~\ref{sec:visually_grounded_model}) exhibit the ME bias. To test this, we present the trained model with two images: one showing familiar and one showing a novel object.
The model is then prompted to identify which image a novel spoken word refers to (Section~\ref{sec:me}).
We denote this ME test as the \me\ test.
With this, we also introduce our notation for specific tests: \texttt{\small <image one type>--<image two type>}, with the type of the audio query underlined. 
The class of the audio query will match the one of the underlined image, unless explicitly stated.
Table~\ref{tab:setup-summary} in Appendix~\ref{append:notation} contains a cheat sheet to understand the tests' notation.

Before we look at our target \me\ ME test, it is essential to ensure that our model has successfully learned to distinguish the familiar classes encountered during training; testing for the ME bias would be premature if the model does not know the familiar classes.
We therefore perform a \familiar\ test, where the task is to match a word query from a familiar class to one of two images containing familiar classes.

Table~\ref{tab:me} presents the results of these two tests for the different \model\ variations described in Section~\ref{sec:initialisations}.
The results of the \familiar\ test show that all the model variations \changed{can distinguish} between familiar classes. 
The vision (AlexNet) initialisation of the vision branch contributes more than the audio (CPC) initialisation\changed{: the two best \familiar\ models both use vision initialisation.}
\changed{Our statistical tests confirm the reported patterns: all model variations are significantly better than the random baseline, and adding the visual (AlexNet) and/or audio (CPC) initialisation to the basic model significantly improves \model's accuracy on the \familiar\ test.}

We now turn to the ME test. The results are given in the \me\ column of Table~\ref{tab:me}.
All \model\ variations exhibit the ME bias, with above-chance accuracy in matching a novel audio segment to a novel image, as also confirmed by our statistical significance test (Appendix~\ref{append:stats}).
From the table, the strongest ME bias is found in the \model\ variation that initialises both the audio (CPC) and vision (AlexNet) branches (row 5), followed by the variation with the vision initialisation alone (row 3).  
Surprisingly, using CPC initialisation alone reduces the strength of the ME bias (row 2 vs row 4). 
Again, these results are confirmed by our statistical tests.
To summarise: even the basic \model\ has the ME bias, but the AlexNet initialisation makes it noticeably stronger.

\begin{table*}[tb]
    \small
      	\newcommand{\cmark}{\textcolor{black}{\ding{51}}}%
        \newcommand{\xmark}{\textcolor{gray!30}{\ding{55}}}%
	\centering
	\renewcommand{\arraystretch}{1.2}
        \newcommand{\ii}[1]{{\footnotesize \textcolor{gray}{#1}}} 
	\begin{tabularx}{\linewidth}
    {@{}r@{\ \ }Lcccccc@{}}
		\toprule
		& & \multicolumn{2}{c}{Model initialisation} & 
		\multicolumn{4}{c}{Accuracy (\%)}\\
		\cmidrule(lr){3-4}
		\cmidrule(l){5-8}
		& & Audio & Vision & \Me & \Novel & \MeOther & \FamiliarWithNovel\\
		\midrule
		\ii{1} & Random baseline  & N/A & N/A & 49.92 & 49.85 & 49.72 & 50.58\\
            \addlinespace
            \ii{2} & \multirow{4}{*}{\model} & \xmark & \xmark & 57.29 & 51.05 & 55.52 & 69.68\\
            \ii{3} &                         & \xmark & \cmark & 59.32 & 48.74 & 58.51 & 86.92\\
            \ii{4} &                         & \cmark & \xmark & 55.92 & 50.52 & 53.41 & 70.93\\
            \ii{5} &                         & \cmark & \cmark & 60.27 & 49.92 & 58.41 & 82.88\\
		\bottomrule
	\end{tabularx}
        \caption{To ensure that the ME bias is real and not because of a peculiarity of our setup, we compare the ME test (\me) to three sanity check experiments for the different variants of \model.}
	\label{tab:sanity}
\end{table*}

\begin{figure}[!t]
	\centering
        \includegraphics[width=\linewidth]{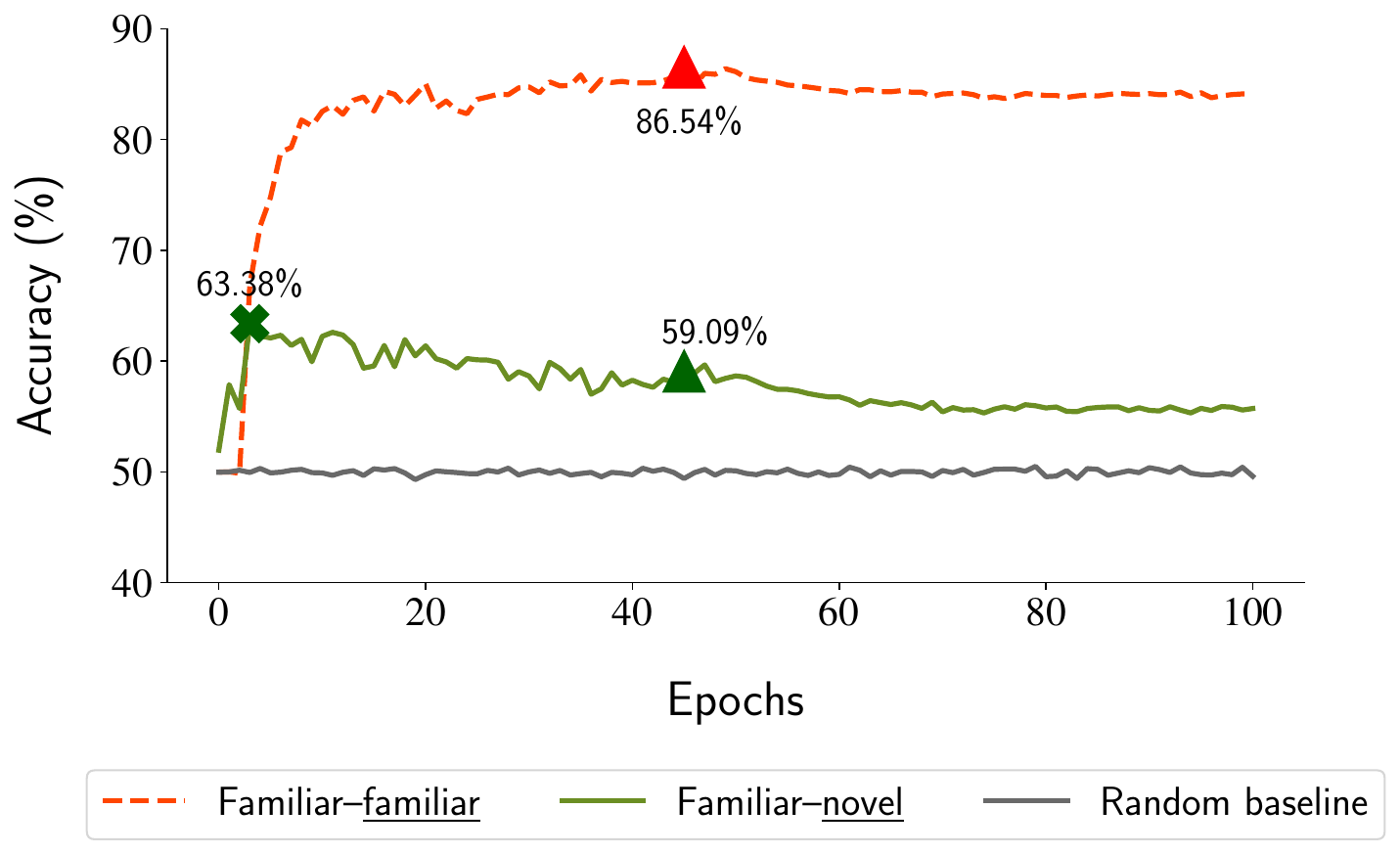}
        \caption{\model's performance over training epochs. The cross indicates the highest overall ME \me\ score. The triangles show the scores at the point where the best \familiar\ score occurs. Results are for the variant of \model\ with both CPC and AlexNet initialisations, and performance is averaged over five training runs.
        }
	\label{fig:smoothed_english_cpc_alexnet}
\end{figure}

To investigate whether the reported accuracies are stable over the course of learning, we consider \model's performance over training epochs on the two tests: \familiar\ and \me. 
We use the model variation with the strongest ME bias, i.e., with both the audio and vision branches initialised. Figure~\ref{fig:smoothed_english_cpc_alexnet} shows that the ME bias (\me, green solid line) is stronger early on in training and then decreases later on.  
The pattern is similar for the \familiar\ score (red dashed line), but the highest score in this case is achieved later in training than the best \me\ score.  
\changed{The scores stabilise after approximately 60 epochs; at this epoch, the model's accuracy is 84.38\% on the \familiar\ task and 56.78\% on the \me\ task (numbers not shown in the figure).}
This suggests that the results reported above for both tests are robust and do not only hold for a particular point in training. 

In summary, we found that a visually grounded speech model, \model, learns the familiar set of classes and thereafter exhibits a consistent and robust ME bias. 
This bias gets stronger when the model is initialised with prior visual knowledge, although the results for the audio initialisation are inconclusive.
Whereas the strength of the ME bias slightly changes as the model learns, it is consistently above chance, suggesting that this is a stable effect in our model.

%% file: further-analysis.tex
\section{Further analyses}
\label{sec:analysis}

We have shown that our visually grounded speech model has the ME bias.
However, we need to make sure that the observed effect is really due to the ME bias and is not a fluke.
In particular, because our model is trained on natural images, additional objects might appear in the background, and there is a small chance that some of these objects are from the novel classes.
As a result, the model may learn something about the novel classes due to information leaking from the training data.
Here we present several sanity-check experiments to show that we observe a small leakage for one model variant, but it does not account for the strong and consistent ME bias reported in the previous section.
Furthermore, we provide additional analyses that show how the model structures its audio and visual representation spaces for the ME bias to emerge.

\subsection{Sanity checks}
\label{subsec:sanity_tests}

The \me\ column in Table~\ref{tab:sanity} repeats the ME results from Section~\ref{sec:results}. 
We now evaluate these ME results against three sanity-check experiments.

We start by testing the following: If indeed the model has a ME bias, it should not make a distinction between two novel classes.
So we present \model\ with two novel images and a novel audio query in a \novel\ test.
Here, one novel image depicts the class referred to by the query, and the other image depicts a different novel class.
If the model does not know the mappings between novel words and novel images, it should randomly choose between the two novel images.
The results for this \novel\ test in Table~\ref{tab:sanity} are close to 50\% for all \model's variations, as expected.

\changed{Surprisingly, our statistical test shows significant differences between the baseline and two out of the four variations: \model\ with full random initialisation scores higher than the baseline on this \novel\ task, and \model\ with the vision initialisation lower. Since the differences between each model and the baseline are small and in different directions (one model scores lower and the other higher), we believe these patterns are not meaningful. At the same time, one possible explanation of the above-chance performance of \model\ with random initialisation is that there may be some leakage of information about the novel classes that may appear in the background of the training images. 
To test whether our ME results can be explained away by this minor leakage, we observe that the model's scores in the \me\ task (the ME task) are noticeably higher than the scores in the \novel\ task. An additional statistical test (Appendix~\ref{append:stats})
shows that the differences between \model's scores across the two tasks are, indeed, statistically significant for three out of the four variations (except the one with the audio initialisation alone).
This suggests that the ME bias cannot be explained away by information leakage for most model variations.}

To further stress test that the model does not reliably distinguish between novel classes, we perform an additional test: \meOther.
In the standard \me\ ME test, the model is presented with a familiar class (e.g., \textsc{elephant}) and a novel class (\textsc{guitar}) and correctly matches the novel query word \textit{guitar} to the novel class.
If the model truly uses a ME bias (and not a mapping between novel classes and novel words that it could potentially infer from the training data), then it should still select the novel image (\textsc{guitar}) even when prompted with a mismatched novel word, say \textit{ball}. Therefore, we construct a test to see whether a novel audio query would still be matched to a novel image even if the novel word does not refer to the class in the novel image.
Results for this \meOther\ test in Table~\ref{tab:sanity}---where the asterisk indicates a mismatch in classes---show that the numbers are very close to those in the standard \me\ ME test. All the \model\ variations therefore exhibit a ME bias: a novel word query belongs to any novel object, even if the two are mismatched, since the familiar object already has a name. 
Our statistical tests support this result.

Finally, in all the results presented above, \model\ has a preference for a novel image. 
One simple explanation that would be consistent with all these results (but would render them trivial) is if the model always chose a novel object when encountering one (regardless of the input query). 
To test this, we again present the model with a familiar and a novel object, but now query it with a familiar word.  
The results for this \familiarWithNovel\ test in Table~\ref{tab:sanity} show that all \model\ variations 
\changed{achieve high scores in selecting the}
familiar object. 
Again, our significance test confirms that all the scores are significantly higher than random.

\subsection{\changed{Why do we see a ME bias?}}
\label{subsec:causes_of_ME}

\begin{figure}[!t]
    \centering
    \includegraphics[width=0.95\linewidth]{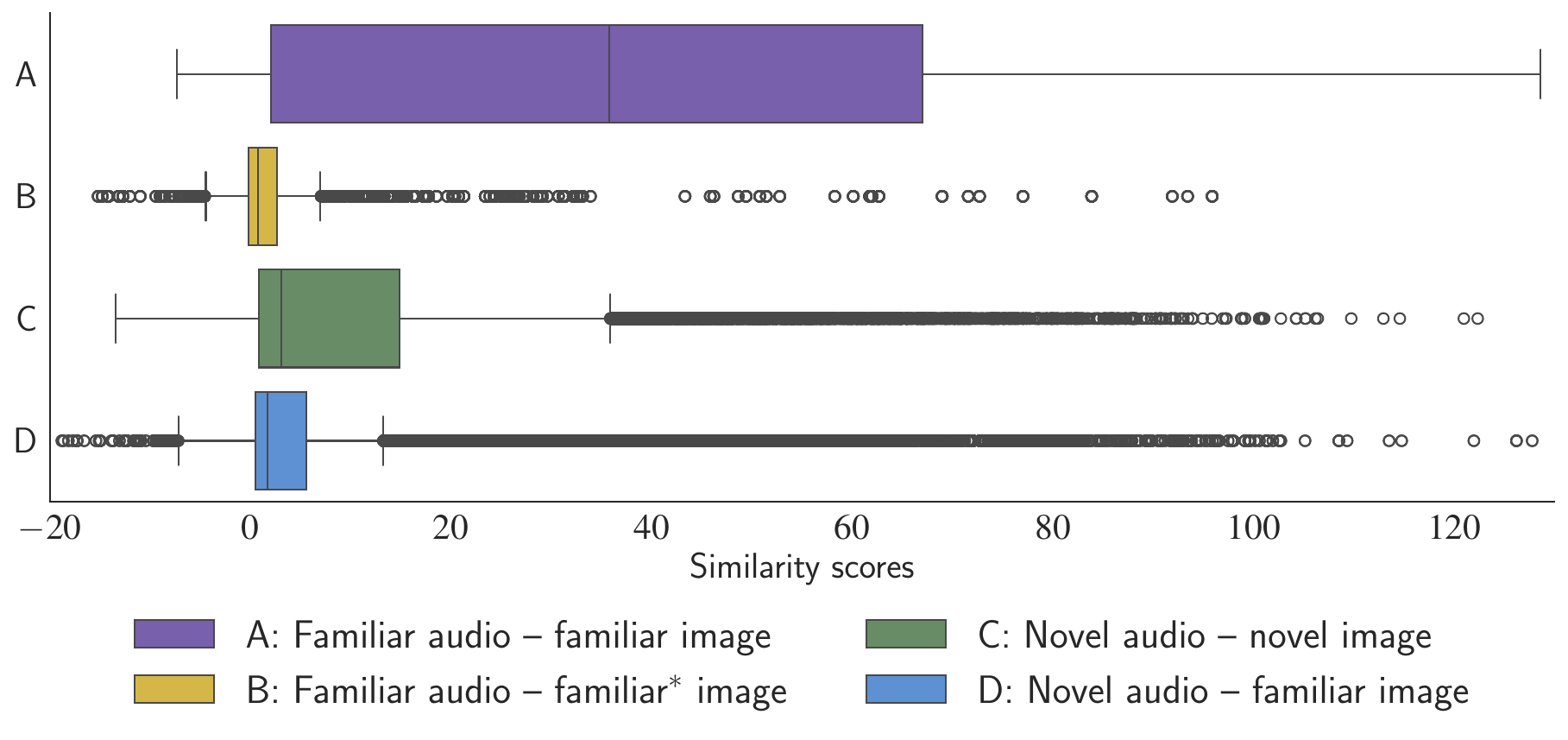}%
    \caption{\changed{
    A box plot of similarities for four types of audio--image comparisons with \model.
    The audio--image examples of a familiar class have higher similarity (A) than mismatched familiar instances (B).
    Novel class instances are in-between (C), but they aren't placed as close as the learned familiar classes~(A). }
    \changed{Novel instances (C) are still closer to each other than to familiar ones~(D).
     }}
    \label{fig:densities}
\end{figure}

We have now established that the \model\ visually grounded speech model exhibits the ME bias.
But this raises the question:
Why does the model select the novel object rather than the familiar one?
How is the representation space organised for this to happen?
We attempt to answer these questions %
by analysing different cross-modal audio--image comparisons made in both the \familiar\ and \me\ (ME) tests.
Results are given in 
\changed{Figure~\ref{fig:densities}},
where we use 
\model\ with both visual and audio encoders initialised (row 5, Table~\ref{tab:sanity}).

First, in the \familiar\ setting we compare two similarities:
(A) the \model\ similarity scores between a familiar audio query and a familiar image from the same class against (B) the similarity between a familiar audio query and a familiar image from a different class (indicated with familiar$^*$).
Perhaps unsurprisingly given the strong \familiar\ performance in Table~\ref{tab:me},
we observe that the similarities of matched pairs (familiar audio -- familiar image, A) 
are substantially higher than the similarities of mismatched pairs (familiar audio -- familiar$^*$ image, B).
This organisation of the model's representation space can be explained by the contrastive objective \changed{in Equation~\ref{eq:loss}}, which ensures that the words and images from the same familiar class are grouped together, and different classes are pushed away from one another.

But where do the novel classes fit in?
To answer this question, we consider two types of comparisons from the \me\ ME setting:
(C) the \model\ similarity scores between a novel query and a novel image (from any novel class) against
(D) the similarity between a novel query and a familiar image.
We observe that the novel audio -- novel image similarities (C) are \changed{typically} higher than the novel audio~-- familiar image similarities (D). 
I.e., novel words in the model's representation space are closer to novel images than to familiar images.
As a result, a novel query on average is closer to \textit{any} novel image than to familiar images, which sheds light on why we observe the ME bias. 

The similarities involving novel words (C and D) are \changed{normally} higher than those of the mismatched familiar classes (B).
This suggests that novel samples are closer to familiar samples than familiar samples from different classes are to one another. 
In other words, during training, the model learns to separate out familiar classes (seen during training), but then places the novel classes (not seen during training) relatively close to at least some of the familiar ones. 
Crucially, samples in the novel regions are still closer to each other \changed{(C)} than they are to any of the familiar classes (as indicated by D).

\begin{figure}[!t]
    \centering
    \includegraphics[width=0.95\linewidth]{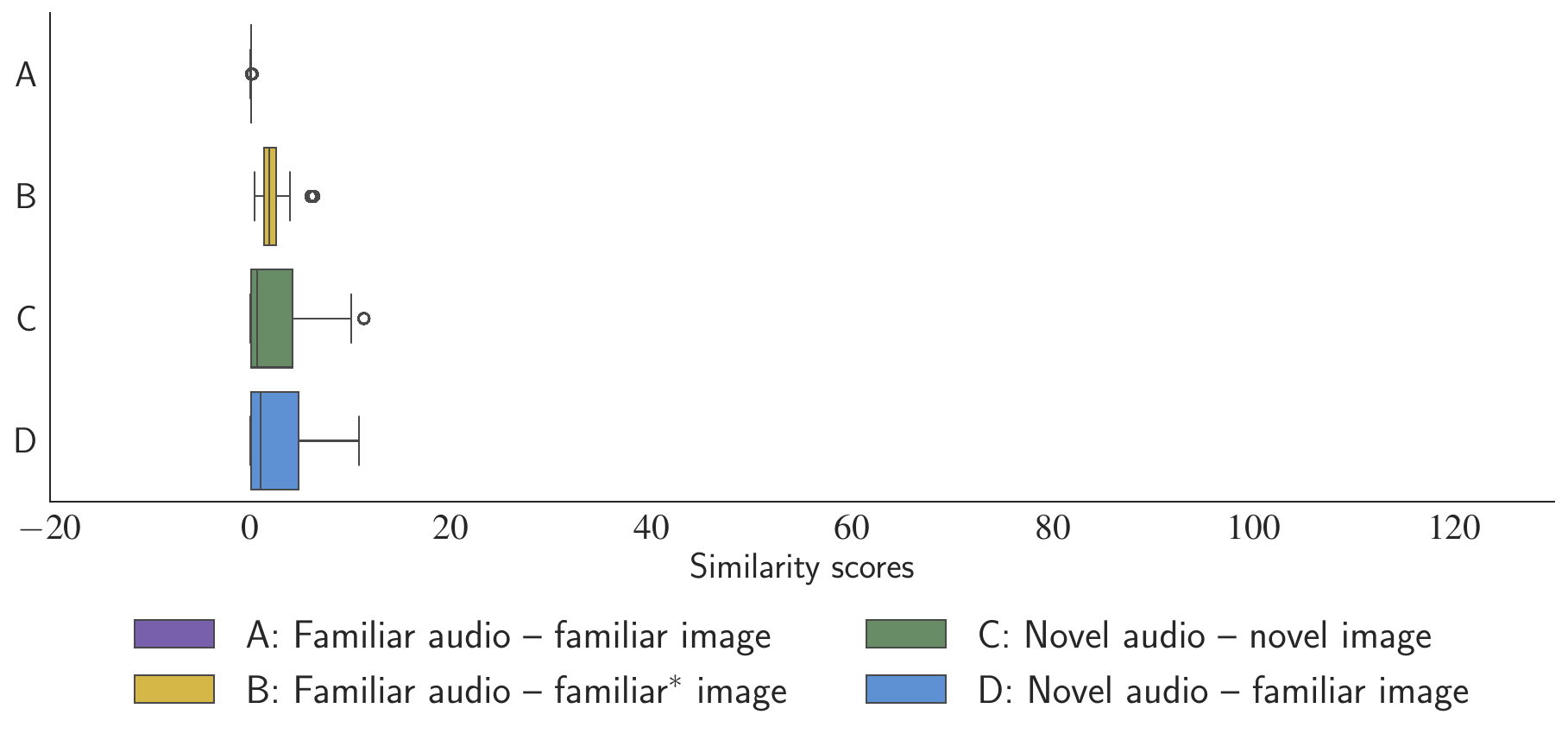}%
    \caption{\changed{
    The same analysis as in Figure~\ref{fig:densities}, but for \model\ before training.
    We can see how similarities are affected through training.
    }} 
    \label{fig:untrained_densities}
\end{figure}

\changed{How does the contrastive loss in Equation~\ref{eq:loss} affect the representations of novel classes during training, given that the model never sees any of these novel instances?
In Figure~\ref{fig:untrained_densities} we plot the same similarities as we did in Figure~\ref{fig:densities} but instead we use the model weights before training.
It is clear how training raises the similarities of matched familiar inputs (A) while keeping the similarities of mismatched familiar inputs low (D), which is exactly what the loss is designed to do. But how are novel instances affected?
One a priori hypothesis might be that training has only a limited effect on the representations from novel classes. But, by comparing Figures~\ref{fig:densities} and~\ref{fig:untrained_densities}, we see that this is not the case: similarities involving novel classes change substantially during training (C and D).
The model thus uses information from the familiar classes that it is exposed to, to update the representation space, affecting both seen and unseen classes.}

\subsection{\changed{Finer-grained analysis}}
\label{subsec:acoustics-vs-me}

\begin{figure*}[!t]
\centering
\subcaptionbox{Mutual exclusivity bias per word. Higher accuracy (dots to the right) indicates a stronger ME bias. \label{fig:anti-me_bias}}%
  [.32\linewidth]{\includegraphics[width=0.30\textwidth]{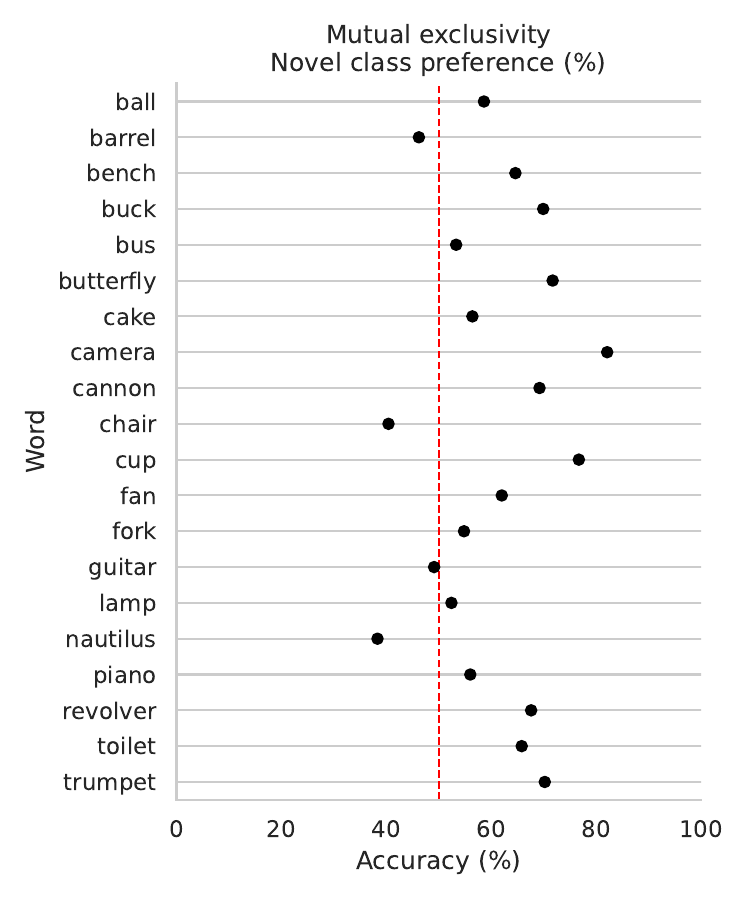}}  
  \hfill
\subcaptionbox{Percentage of times a familiar image is selected for a novel audio.
\changed{%
Some entries are
empty because these were never compared in any of the sampled
episodes.}
\label{fig:anti-me_bias_confusion_matrix}}
  [.32\linewidth]{\includegraphics[width=0.30\textwidth]{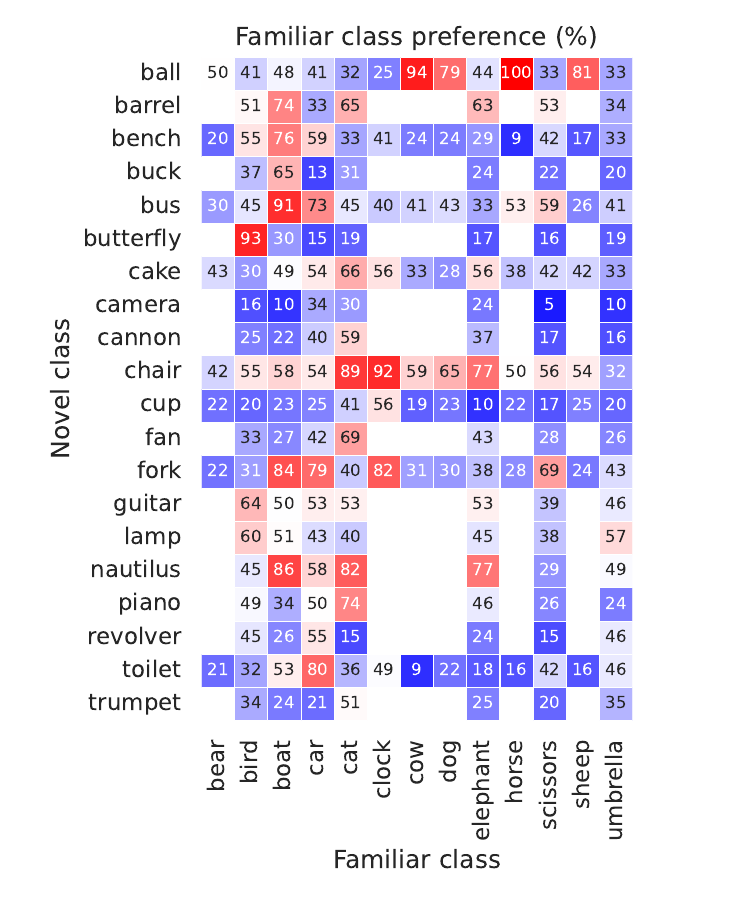}}  
  \hfill
\subcaptionbox{Similarities of audio embeddings between novel and familiar words. The numbers are cosine similarity times 100. Lighter shades are associated with higher similarity. \label{fig:anti-me_bias_audio_confusion_matrix}}  
  [.32\linewidth]{\includegraphics[width=0.30\textwidth]{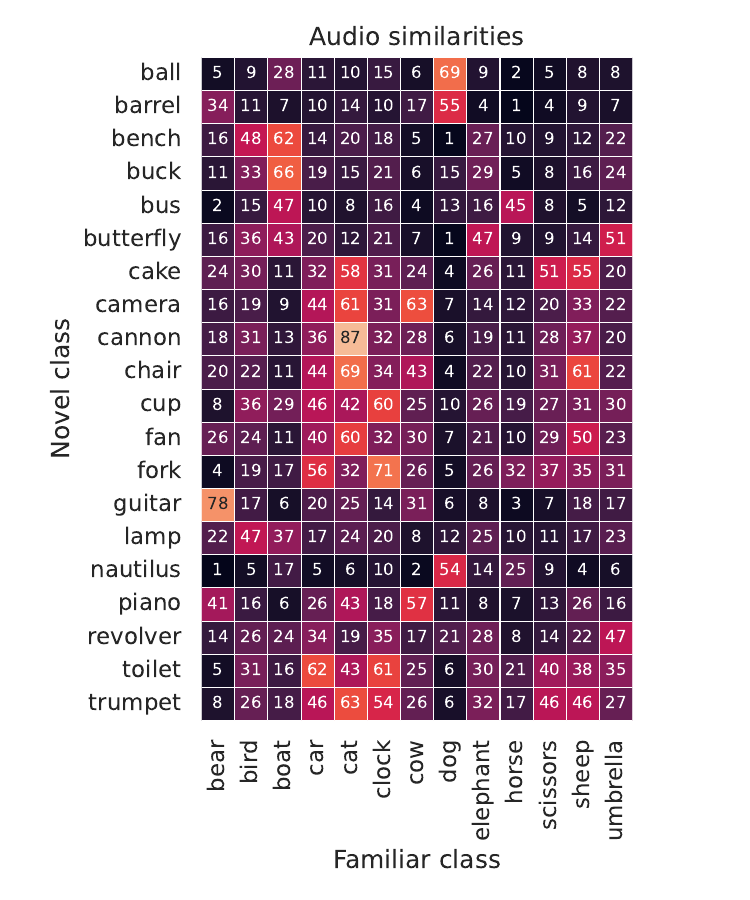}} 
    \vspace*{-5pt}
    \caption{\changed{A finer-grained analysis looking at the ME bias individually for each of the 20 novel words.}}
    \label{fig:me-vs-sim}
\end{figure*}

\changed{We have seen a robust ME bias in the aggregated results above. But what do results look like at a finer level?}
\changed{We now consider each of the 20 novel words individually and}
compute how often the model selects the corresponding novel image (Figure~\ref{fig:anti-me_bias}) or any of the familiar images (Figure~\ref{fig:anti-me_bias_confusion_matrix}).
While most of the novel words 
are associated with the
ME bias (Figure~\ref{fig:anti-me_bias}, dots to the right of the vertical red line), a small number of words yield a strong anti-ME bias when paired with certain familiar words (Figure~\ref{fig:anti-me_bias_confusion_matrix}, red cells).
\changed{E.g., for the novel word \word{bus}, in 91\% of the test cases the model picks an image of a familiar class \image{boat} rather than an image of the novel class \image{bus}.}
\changed{It is worth emphasising that the ME bias isn't absolute: even in human participants it isn't seen in 100\% of test cases.
Nevertheless, it is worth investigating why there is an anti-ME bias for some particular words (something that is easier to do in a computational study compared to human experiments).}

\changed{One reason for an anti-ME result is the phonetic similarity of a novel word to familiar words.}
\changed{E.g., \word{bus} and \word{boat} start with the same consonant followed by a vowel.
If we look at Figure~\ref{fig:anti-me_bias_audio_confusion_matrix}, which shows the cosine similarities between the learned audio embeddings from \model, we see that spoken instances of \word{bus} and \word{boat} indeed have high similarity. In fact, several word pairs starting with the same consonant (followed by a vowel) have high learned audio similarities, e.g., \word{buck}--\word{boat}, \word{bench}--\word{boat} and \word{cake}--\word{cat}, all translating to an anti-ME bias in Figure~\ref{fig:anti-me_bias_confusion_matrix}.}

\changed{However, the anti-ME bias cannot be explained by
acoustic similarity alone: some anti-ME pairs have low audio similarities, e.g., \word{nautilus}--\word{elephant}.
For such cases, the representation space must be structured differently from the aggregated analysis in Section~\ref{subsec:causes_of_ME} (otherwise we would see a ME bias for these pairs).
Either the spoken or the visual representation of a particular class can be responsible (or both).  To illustrate this, we zoom in on the two novel words showing the strongest anti-ME results in Figure~\ref{fig:anti-me_bias}: \word{nautilus} and \word{chair}.}

\begin{figure}[tb]
     \centering
     \begin{subfigure}[b]{\linewidth}
         \centering
         \includegraphics[width=\linewidth]{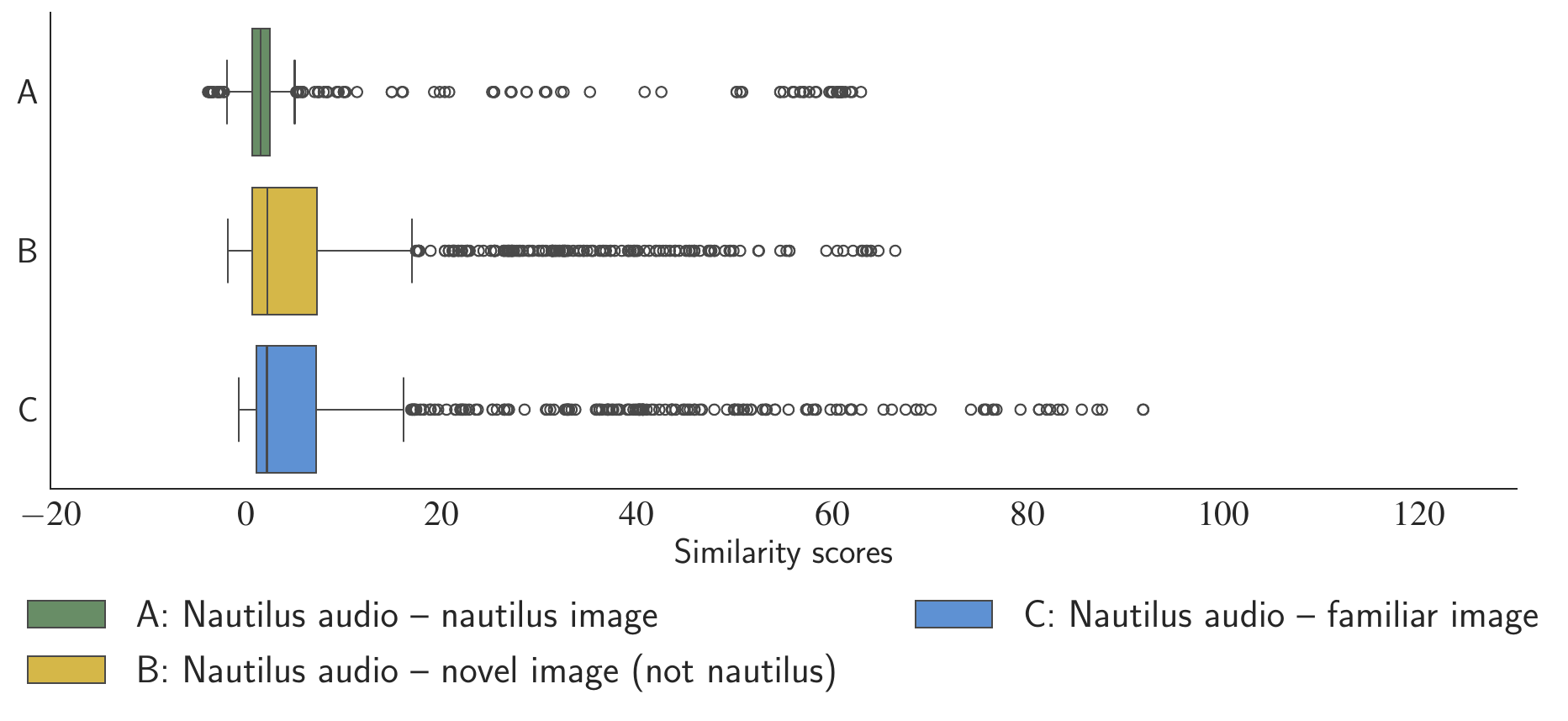}%
         \caption{
         \changed{Similarity scores for comparisons involving \word{nautilus}.}
         }
         \label{fig:nautilus}
     \end{subfigure}
     ~\\[-10pt]
     \begin{subfigure}[b]{\linewidth}
         \centering
         \includegraphics[width=\linewidth]{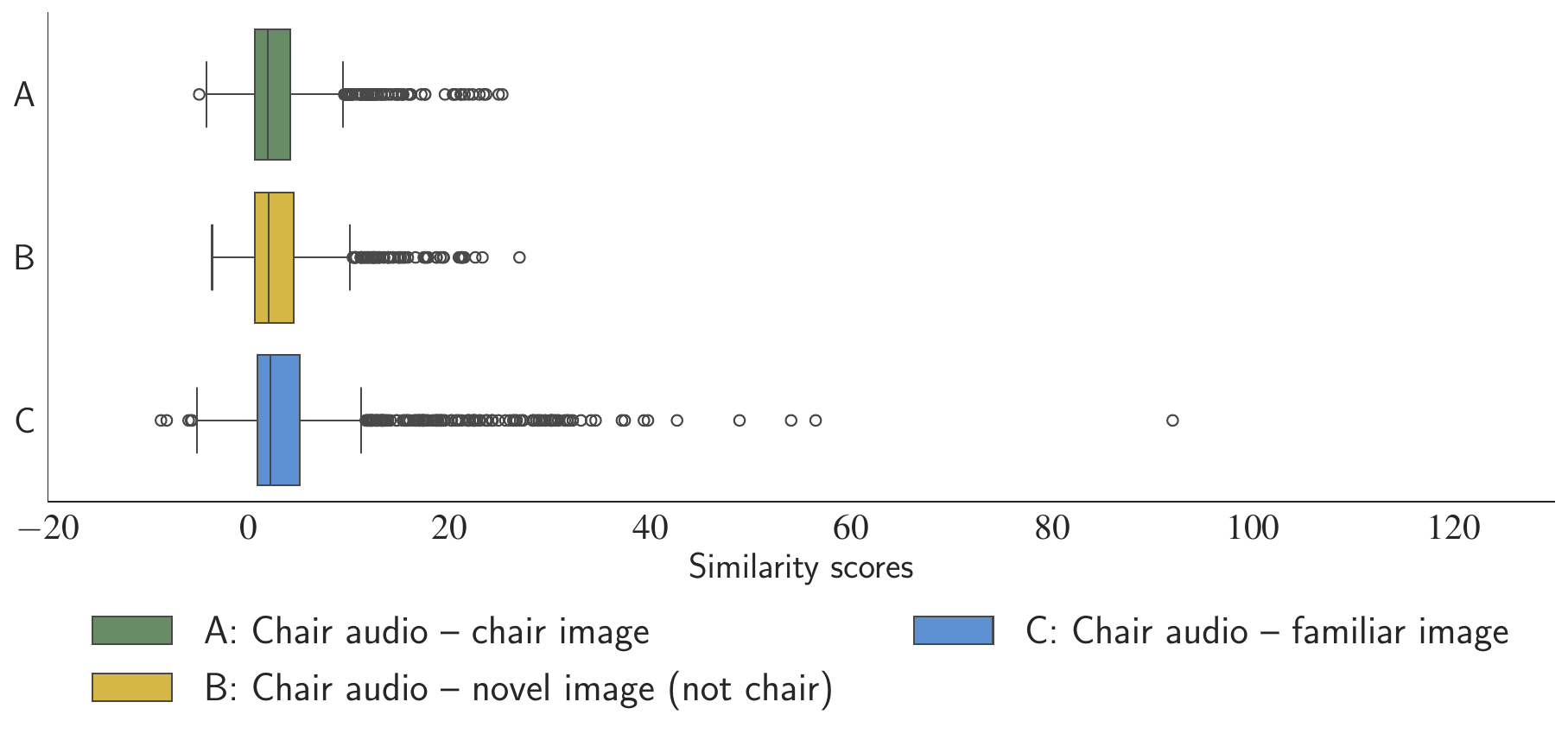}%
         \caption{
         \changed{Similarity scores for comparisons involving \word{chair}.}
         }
         \label{fig:chair}
     \end{subfigure}
     \label{fig:per_keyword_densities}
     \vspace*{-15pt}
     \caption{\changed{
        Box plots of similarities between combinations of novel and familiar class instances focussed on two classes: (a) \word{nautilus} and (b) \word{chair}.} }
\end{figure}

\changed{
Figure~\ref{fig:nautilus} presents a similar analysis to that of Figure~\ref{fig:densities} but specifically for \word{nautilus}.
We see the anti-ME bias: \word{nautilus} audio is more similar to familiar images (C) than to \image{nautilus} images (A).
This is the reverse of the trend in Figure~\ref{fig:densities} (C vs D).
Is this due to the \word{nautilus} word queries or the \image{nautilus} images?
Here in Figure~\ref{fig:nautilus}, box B shows what happens when we substitute the \image{nautilus} images from box A with any other novel image: the similarity goes up.
This means that \image{nautilus} images 
are not placed in the same area of the representation space as the other novel images.
But this isn't all: boxes B and C are also close to each other. Concretely, if we compare B vs C here in Figure~\ref{fig:nautilus} to C vs D in Figure~\ref{fig:densities}, then we still do not see the difference corresponding to the ME result, as in the latter case. This means that the
\image{nautilus} audio is also partially responsible for the anti-ME result here in that it is placed close to familiar images.}

\changed{Let us do a similar analysis for \word{chair}: Figure~\ref{fig:chair}.
We again see the anti-ME results by comparing A and C.
But now swapping out \image{chair} images for other novel images (B) does not change the similarities.
In this case, the culprit is therefore mainly the \word{chair} audio.
}

\changed{Further similar analyses can be done to look at other anomalous cases. But it is worth noting, again, that the aggregated ME scores from Section~\ref{sec:results} are typically between 55\% and 61\% (not 100\%). So we should expect some anti-ME trends in some cases, and the analysis in this section shows how we can shed light on those.}

\subsection{\changed{How specific are our findings to \model?}}
\label{sec:general_experiments}

\changed{We have considered one visually grounded speech model, namely \model.
How specific are our findings to this particular model?
While several parts of our model can be changed to see what impact they have, we limit our investigation to two potentially important components:
the loss function and the visual network initialisation.
}

\paragraph{\changed{Loss function.}}
\changed{Apart from the loss in Equation~\ref{eq:loss}, we now look at two other contrastive losses.
The hinge loss is popular in many visually grounded speech models \cite{harwath_unsupervised_2016, chrupala_representations_2017}.}
\changed{
It uses a piece-wise linear function to ensure a greater similarity for matched pairs:
\begin{equation}
\small
\begin{aligned}
\ell = & \sum_{i=1}^{N_\textrm{neg}} \textrm{max}\left(0, S(\ab_i^{-}, \vb) - S(\ab, \vb) + m\right)\\
&+ \sum_{i=1}^{N_\textrm{neg}}\textrm{max}\left(0, S(\ab, \vb_i^{-}) - S(\ab, \vb) + m\right)
\end{aligned}
\label{eq:hinge}
\end{equation}
where $m = 1$ is a margin parameter.
We sample negatives within a batch, similar to \citet{harwath_unsupervised_2016}.}

\changed{InfoNCE is a loss typically employed by self-supervised models~\cite{oord_representation_2019} and vision--text models \cite{jia_scaling_2021, li_align_2021, radford_learning_2021}.}
\changed{
It uses the logistic function to select a positive pair from among a set of negatives:
\begin{equation}
\small
\begin{aligned}
    \ell = & \log \frac{\exp S(\ab, \vb)}{\exp S(\ab, \vb) + \sum_{i=1}^{N_\textrm{neg}}{\exp S(\ab, \vb_i^{-})}} \\
         &+ \log \frac{\exp S(\ab, \vb)}{\exp S(\ab, \vb) + \sum_{i=1}^{N_\textrm{neg}}{\exp S(\ab_i^{-}, \vb)}}
\end{aligned}
\label{eq:infonce}
\end{equation}}

\changed{Apart from changing the loss, the rest of the \model\ structure is retained.
Results are shown in Table~\ref{tab:contrastive_losses} for models that use self-supervised CPC and AlexNet initialisations.
The two new losses can learn the familiar classes and exhibit a ME bias.
In fact, an even better \familiar\ performance and a stronger ME bias (\me) are obtained with the InfoNCE loss.}
\changed{This loss should therefore be considered in future work studying the ME bias in visually grounded speech models.}

\begin{table}[!bt]
        \small
	\newcommand{\cmark}{\color{black}\ding{51}}%
        \newcommand{\xmark}{\color{gray!30}\ding{55}}%
	\centering
	\renewcommand{\arraystretch}{1.2}
        \newcommand{\ii}[1]{{\footnotesize \textcolor{gray}{#1}}} 
	\begin{tabularx}{\linewidth}
    {@{}Lcc@{}}
    \toprule 
    & \multicolumn{2}{c}{Accuracy (\%)}\\
    \cmidrule(l){2-3}
    Loss & {\footnotesize \Familiar} & {\footnotesize \Me}\\
    \midrule
    \model~\eqref{eq:loss}     & 83.20 & 60.27\\
    Hinge~\eqref{eq:hinge}     & 87.21 & 57.85\\
    InfoNCE~\eqref{eq:infonce} & 93.16 & 63.91\\
    \bottomrule
    \end{tabularx}
    \caption{
    \changed{The effect of different losses on the ME test (\me) and the sanity check (\familiar).}
    }
    \label{tab:contrastive_losses}
\end{table}

\begin{table}[!bt]
    \small
    \newcommand{\cmark}{\color{black}\ding{51}}%
    \newcommand{\xmark}{\color{gray!30}\ding{55}}%
    \centering
    \renewcommand{\arraystretch}{1.2}
    \newcommand{\ii}[1]{{\footnotesize \textcolor{gray}{#1}}} 
    \begin{tabularx}{\linewidth}
    {@{}Lcc@{}}
    \toprule
    & \multicolumn{2}{c}{Accuracy (\%)}\\
    \cmidrule(l){2-3}
    Vision initialisation & {\footnotesize\Familiar} &{\footnotesize\Me}\\
    \midrule
    Self-supervised & 83.20 & 60.27\\
    Supervised & 87.08 & 61.66 \\
    \bottomrule
    \end{tabularx}
    \caption{
    \changed{The effect of using a self-supervised or supervised version of AlexNet for visual initialisation. Scores for the ME test (\me) and the sanity check (\familiar) are reported.}
    }
    \label{tab:alexnet}
\end{table}

\paragraph{\changed{Visual network initialisation.}}
\changed{In Section~\ref{sec:results}
we saw that vision initialisation contributes most to the ME strength.
Here we investigate whether we can get an even greater performance boost if we initialise \model\ using a supervised version of AlexNet instead of the self-supervised variant used thus far.
Both the self-supervised~\cite{koohpayegani_compress_2020} and supervised~\cite{krizhevsky_imagenet_2017} versions of AlexNet are trained on ImageNet~\cite{deng_imagenet_2009}, so
we can fairly compare \model\ when initialised with either option.
}
\changed{
Both \model\ variants shown in Table~\ref{tab:alexnet} make use of CPC initialisation.
We observe that the supervised AlexNet initialisation performs better on the \familiar\ task than the self-supervised initialisation.
However, the ME (\me) results with the supervised AlexNet initialisation are only slightly higher than with the self-supervised initialisation.
}

\changed{While there is a broad space of visually grounded models that could be used to consider the ME task, it is encouraging that all the variants in this work show the bias.}

%% file: conclusion.tex
\section{Conclusion and future work}

Mutual exclusivity (ME) is a constraint employed by children learning new words: a novel word is assumed to belong to an unfamiliar object rather than a familiar one.
In this study, we have demonstrated that a representative visually grounded speech model exhibits a consistent and robust ME bias, similar to the one observed in children. We achieved this by training the model on a set of spoken words and images and then asking it to match a novel acoustic word query to an image depicting either a familiar or a novel object.
We considered different initialisation approaches simulating prior language and visual processing abilities of a learner.
The ME bias was observed in all cases, with the strongest bias occurring when more prior knowledge was used in the model (initialising the vision branch had a particularly strong effect).

In further analyses we showed that the ME bias is strongest earlier on in model training and then stabilises over time. 
In a series of additional sanity-check tests we showed that the ME bias was not an artefact: it could not be explained away by possible information leakage from the training data or by trivial model behaviours. 
We found that the resulting embedding space is organised such that novel classes are mapped to a region distinct from the one containing familiar classes, and that different familiar classes are spread out over the space to maximise the distance between familiar classes. 
As a result, novel words are mapped on to novel images, leading to a ME bias. 
\changed{Lastly, we showed that the ME bias is robust to model design choices in experiments where we changed the loss function and used a supervised instead of self-supervised visual initialisation approach.}

\changed{Future work can consider whether using a larger number of novel and familiar classes affects the results. Another interesting avenue for future studies resolves around multilingualism.}
Following on from the original ME studies 
with young children, \citet{byers-heinlein_monolingual_2009} and \citet{kalashnikova_effects_2015}, among others, have looked at how multilingualism affects the use of the ME constraint. 
This setting is interesting since in the multilingual case different words from the distinct languages are used to name the same object.
These studies showed that in bi- and trilingual children from the same age group, the ME bias is not as strong as in monolingual children.
We plan to investigate this computationally in future work. 

\section*{Acknowledgements}

This work was supported through a Google DeepMind scholarship for LN and a research grant from Fab~Inc.\ for HK.
DO 
was partly supported by the European Union’s HORIZON-CL4-2021-HUMAN-01 research and innovation programme under grant agreement no.\ 101070190 AI4Trust.
We would like to thank Benjamin van Niekerk for useful discussions about the analysis.
We would also like to thank the anonymous reviewers and action editor for their valuable feedback.

%% file: more_results.tex
\section{Testing for statistical significance}
\label{append:stats}

To determine whether the differences between our model variations and a random baseline are statistically significant, we fit two types of logistic mixed-effects regression models to the data, where each of them predicts the (binary) model's choice for each test episode. All models are fitted using the lme4 package \citep{bates_fitting_2015}. \changed{Unlike many other statistical tests, mixed-effects models take into account the structure of the data: e.g., certain classes or even individual images/queries are used in multiple pairwise comparisons.}

The first mixed-effects model tests whether each \model's variation is better than the random baseline: it uses the \model\ variation as a predictor variable and random intercepts over trials, test episodes, the specific acoustic realisation of the test query, individual image classes and their pairwise combinations, and specific images in the test episode.

The second mixed-effects model does not consider the random baseline, and instead tests whether adding the visual initialisation, the audio initialisation or a combination of both improves \model: it uses the presence (or lack of) visual initialisation and audio initialisation as two binary independent variables, as well as their interaction, and the same random intercepts as described above. 

\changed{In Section~\ref{subsec:sanity_tests} we additionally test whether \model's scores in the \me\ test are significantly higher than in the \novel\ test. For this, we fit a logistic mixed-effects model to \model's combined scores from both tests,} with test type and model variation as predictor variables, together with their interaction, as well as random intercepts as described above.

\newpage
\section{Test notation}
\label{append:notation}

\begin{table}[!h]
    \small
    \caption{%
    A summary of the evaluation setups in terms of the input types (familiar or novel) used for the audio query and the two images.
    The asterisk indicates different classes for the same input type.
    E.g., \image{familiar} and \image{familiar}$^*$ are two different familiar classes.}
    \label{tab:setup-summary}
    \setlength{\tabcolsep}{4pt}
    \begin{tabularx}{\linewidth}{@{}lCCCC@{}}
        \toprule
        Setup & Query audio & Target image & Other image \\
        \midrule
        \Familiar & \word{familiar} & \image{familiar}  & \image{familiar}$^*$ \\
        \FamiliarWithNovel & \word{familiar} & \image{familiar} & \image{novel} \\
        \midrule
        \Me & \word{novel} & \image{novel}  & \image{familiar} \\
        \Novel & \word{novel} & \image{novel} & \image{novel}$^*$ \\
        \MeOther & \word{novel} & \image{novel}$^*$ & \image{familiar}  \\
        \bottomrule
    \end{tabularx}
\end{table}